\documentclass[sigconf]{acmart}
\AtBeginDocument{%
  }

\copyrightyear{2026}
\acmYear{2026}
\setcopyright{cc}
\setcctype{by}
\acmConference[MM '26] {Proceedings of the 34th ACM International Conference on Multimedia}{November 10--14, 2026}{Rio de Janeiro, Brazil.}
\acmBooktitle{Proceedings of the 34th ACM International Conference on Multimedia (MM '26), November 10--14, 2026, Rio de Janeiro, Brazil}
\acmISBN{979-8-4007-2213-4/2026/11}
\acmDOI{10.1145/3767308.3835434}
\usepackage{balance}  
\usepackage{booktabs, multirow, xcolor, colortbl}
\usepackage{tabularx}
\usepackage{graphicx}
\usepackage{multirow}
\usepackage{threeparttable}
\usepackage{xspace}

\usepackage{enumitem}
\usepackage{color}
\usepackage{colortbl}
\usepackage{bm}
\definecolor{Gray}{gray}{0.9}
\usepackage{booktabs,siunitx}
\usepackage{adjustbox}
\usepackage{makecell}
\usepackage[capitalise]{cleveref}
\begin{document}

\title{Hypergraph-based Multimodal Retrieval-Augmented Generation with Incremental Refinement}

\author{Shenao Chen}
\authornote{Both authors contributed equally to this research.}
\email{241080005@hdu.edu.cn}
\affiliation{%
  \institution{Hangzhou Dianzi University}
  \city{Hangzhou}
  \country{China}
}

\author{Yidan Xu}
\authornotemark[1]
\email{yidanxu2024@163.com}
\affiliation{%
  \institution{Hangzhou Dianzi University}
  \city{Hangzhou}
  \country{China}
}

\author{Xiangmin Han}
\correspondingauthor
\email{hanxiangmin@bjut.edu.cn}
\affiliation{%
  \institution{Beijing University of Technology}
  \city{Beijing}
  \country{China}
}

\author{Rundong Xue}
\email{xuerundong2002@gmail.com}
\affiliation{%
  \institution{Xi'an Jiaotong University}
  \city{Xi'an}
  \country{China}}

\author{Duanpo Wu}
\email{wuduanpo@hdu.edu.cn}
\affiliation{%
  \institution{Hangzhou Dianzi University}
  \city{Hangzhou}
  \country{China}
}

\author{Yuhan Gao}
\correspondingauthor
\email{yuhangao@hdu.edu.cn}
\affiliation{%
  \institution{Hangzhou Dianzi University}
  \city{Hangzhou}
  \country{China}
}

\author{Chenggang Yan}
\email{cgyan@hdu.edu.cn}
\affiliation{%
  \institution{Hangzhou Dianzi University}
  \city{Hangzhou}
  \country{China}
}

\author{Yue Gao}
\email{gaoyue@tsinghua.edu.cn}
\affiliation{%
  \institution{Tsinghua University}
  \city{Beijing}
  \country{China}
}

\renewcommand{\shortauthors}{Shenao Chen et al.}

\begin{abstract}
Modern Multimodal Retrieval-Augmented Generation (M-RAG) systems are fundamentally limited by the binary connectivity paradigm of traditional simple graphs, which fails to capture the intricate, high-order correlations among heterogeneous entities—such as the N-ary relationships between a visual chart, its scattered textual descriptions, and underlying numerical data. Furthermore, existing refinement strategies often rely on exhaustive, full-page reconstruction to align cross-modal information, leading to prohibitive computational redundancy and the introduction of contextual noise in long-form document processing.
In this paper, we propose Hyper-M2RAG, a novel framework that redefines multimodal document retrieval through High-order Hypergraph Representation Learning. We first formalize the document structure as a Multimodal Hypergraph, utilizing hyperedges as unified semantic containers to encapsulate multi-way associations across text, images, and tables, thereby transcending point-to-point modeling. To mitigate semantic fragmentation caused by physical pagination, we introduce an Anchor-driven Incremental Refinement mechanism. Rather than performing a global sweep, our approach surgically identifies boundary-crossing anchors—nodes and reconstructs their local hyper-topology using one-hop neighborhood contexts. This targeted refinement effectively bridges cross-page knowledge gaps with minimal computational footprints.
Extensive evaluations on multimodal benchmarking datasets demonstrate that Hyper-M2RAG significantly outperforms state-of-the-art methods in both retrieval precision and generation coherence. Our code is accessible at \url{https://github.com/ShenAoChen2001/MMHRAG}.
\end{abstract}






\maketitle

\section{Introduction}
\label{sec:intro}
Document Question Answering (Doc-QA) \cite{wang2024leave, wu2025doc} aims to generate accurate answers grounded in document content. With the rapid progress of Large Vision-Language Models (LVLMs), Doc-QA has evolved from text-only understanding to multimodal reasoning over long and structurally complex documents. In real-world scenarios such as academic papers, technical reports, and business documents, critical evidence is often distributed across text, figures, tables, formulas, and page layouts, and may span many pages. This setting requires not only long-context modeling, but also faithful preservation of cross-modal structure and document-level semantic dependencies.

Retrieval-augmented generation (RAG) has become an effective paradigm for document understanding by retrieving relevant evidence before answer generation. Existing RAG frameworks, however, are still largely rooted in text-centric assumptions \cite{wu2025retrievalaugmentedgenerationnaturallanguage, hindi2025enhancing}. Even recent multimodal variants often remain limited in how they represent document structure. A common strategy is to linearize visual content into captions or OCR-derived text, and then perform retrieval over textual surrogates \cite{yu2024visrag, faysse2024colpali}. Although practical, this design inevitably weakens the original spatial organization of the document and discards part of the visual semantics embedded in figures, tables, and layouts. As a result, evidence that depends on visual context, layout proximity, or joint interpretation across modalities can be overlooked during retrieval.

Beyond this general limitation, a more fundamental issue lies in the relational assumption inherited by many graph-based or knowledge-enhanced RAG systems. In text-centric settings, knowledge is commonly modeled through binary relations, typically represented as triples of the form \textit{(head, relation, tail)}. This abstraction is often adequate when semantic dependencies can be decomposed into pairwise links between entities or text spans. However, multimodal documents exhibit a different structural nature. Key evidence is frequently formed not by a single pairwise connection, but by the joint interaction among multiple heterogeneous elements, such as a figure region, its caption, a nearby paragraph, a table cell, and their shared layout context. In such cases, the meaning emerges from a group-wise association rather than from isolated binary edges.

This mismatch makes conventional graph modeling inherently limited for multimodal RAG. When high-order multimodal dependencies are decomposed into a set of pairwise edges, the original evidence unit becomes fragmented. The system may preserve local connections, yet lose the higher-level semantic coherence that binds multiple elements into a single interpretable structure. For example, a textual claim may only be fully supported when a chart, its legend, and the surrounding explanatory paragraph are considered together; reducing this evidence to independent pairwise links weakens structural fidelity and makes downstream reasoning less reliable. Therefore, the challenge in multimodal RAG is not merely to add more modalities into existing retrieval pipelines, but to redesign the underlying relational structure from graph-based pairwise modeling to hypergraph-based high-order organization.

Recent advances in document parsing pipelines \cite{ke2025large}, such as MinerU2.5 \cite{niu2025mineru2}, DeepSeekOCR \cite{wei2026deepseek}, and PaddleOCR \cite{cui2025paddleocr}, make this shift increasingly practical. These systems can extract high-fidelity multimodal elements from complex PDFs, including text blocks, figure regions, table structures, and layout coordinates, thereby providing a stronger foundation for structured multimodal indexing. This progress opens the door to a retrieval framework that operates directly on rich document elements and their structural relations, instead of relying on lossy textual abstractions alone.

Motivated by this gap, we ask the following question: \emph{how can we extend RAG from graph-based pairwise retrieval to hypergraph-based high-order multimodal reasoning, while keeping refinement efficient for long documents?}

To answer this question, we identify two core challenges. \textbf{Challenge 1: Pairwise graph modeling is insufficient for high-order multimodal evidence.} Existing graph-based retrieval structures are designed for binary relations and therefore struggle to preserve group-wise dependencies among text, figures, tables, and layout context. This limitation leads to fragmented evidence representation and weakens the retriever's ability to discover visually grounded or structurally coupled information. \textbf{Challenge 2: Efficient refinement in long multimodal documents remains difficult.} In long documents, relevant evidence and entity relations often span many pages. Existing refinement strategies frequently revisit entire pages or large contexts to recover missing cross-page connections \cite{hsiao2025megarag}, which introduces substantial computational redundancy, especially when multimodal tokens dominate the cost.

We propose \textbf{Hyper-M2RAG}, a hypergraph-based framework for multimodal retrieval-augmented generation. The central idea is to move the structural foundation of RAG from graphs to hypergraphs. Hyper-M2RAG constructs a multimodal hypergraph in which vertices represent textual, visual, and tabular entities, while hyperedges connect multiple heterogeneous elements that jointly express a semantic unit. This design allows the retriever to preserve and exploit high-order multimodal dependencies in a unified topological space, rather than approximating them through decomposed pairwise links. As a result, retrieval becomes natively aware of textual semantics, visual evidence, and layout structure.

To further support long-document reasoning, Hyper-M2RAG introduces an \emph{anchor-driven surgical refinement} mechanism. Our key observation is that entities recurring across pages often signal incomplete or ambiguous structural associations. Instead of reprocessing entire pages, the system identifies these boundary-crossing entities as anchors and selectively reconstructs their local one-hop neighborhoods. By refining only the hyperedges around these anchors and merging the updates back into the global hypergraph, our method concentrates computation on the structurally uncertain regions that matter most. This makes refinement substantially more efficient while preserving document-level reasoning capacity.

Our contributions are summarized as follows:
\begin{itemize}[leftmargin=*, noitemsep, topsep=2pt]
    \item \textbf{From graph to hypergraph for multimodal RAG.} We show that the pairwise relational assumption underlying conventional graph-based RAG is insufficient for multimodal documents, and introduce a hypergraph formulation that naturally models high-order dependencies among text, figures, tables, and layout elements.
    
    \item \textbf{Native multimodal hypergraph indexing.} We leverage visual and layout information during document parsing 
    to extract structured entities and relationships. These extracted 
    textual representations are then organized into a unified hypergraph, 
    supporting retrieval over semantically enriched structural evidence.
    
    \item \textbf{Anchor-driven surgical refinement.} We propose a localized refinement strategy that updates only the structurally ambiguous regions around cross-page anchors, avoiding exhaustive page-level reprocessing and improving efficiency in long-document settings.
    
    \item \textbf{Effective and efficient multimodal reasoning.} Through extensive experiments, we show that Hyper-M2RAG improves multimodal retrieval and downstream answer quality while achieving a better efficiency profile on long and complex documents.
\end{itemize}

The remainder of this paper details the proposed method. Section~\ref{sec:method} formalizes Hyper-M2RAG and presents its multimodal hypergraph construction and anchor-driven refinement mechanism.

\begin{figure*}[t]
    \centering
    \includegraphics[width=\textwidth]{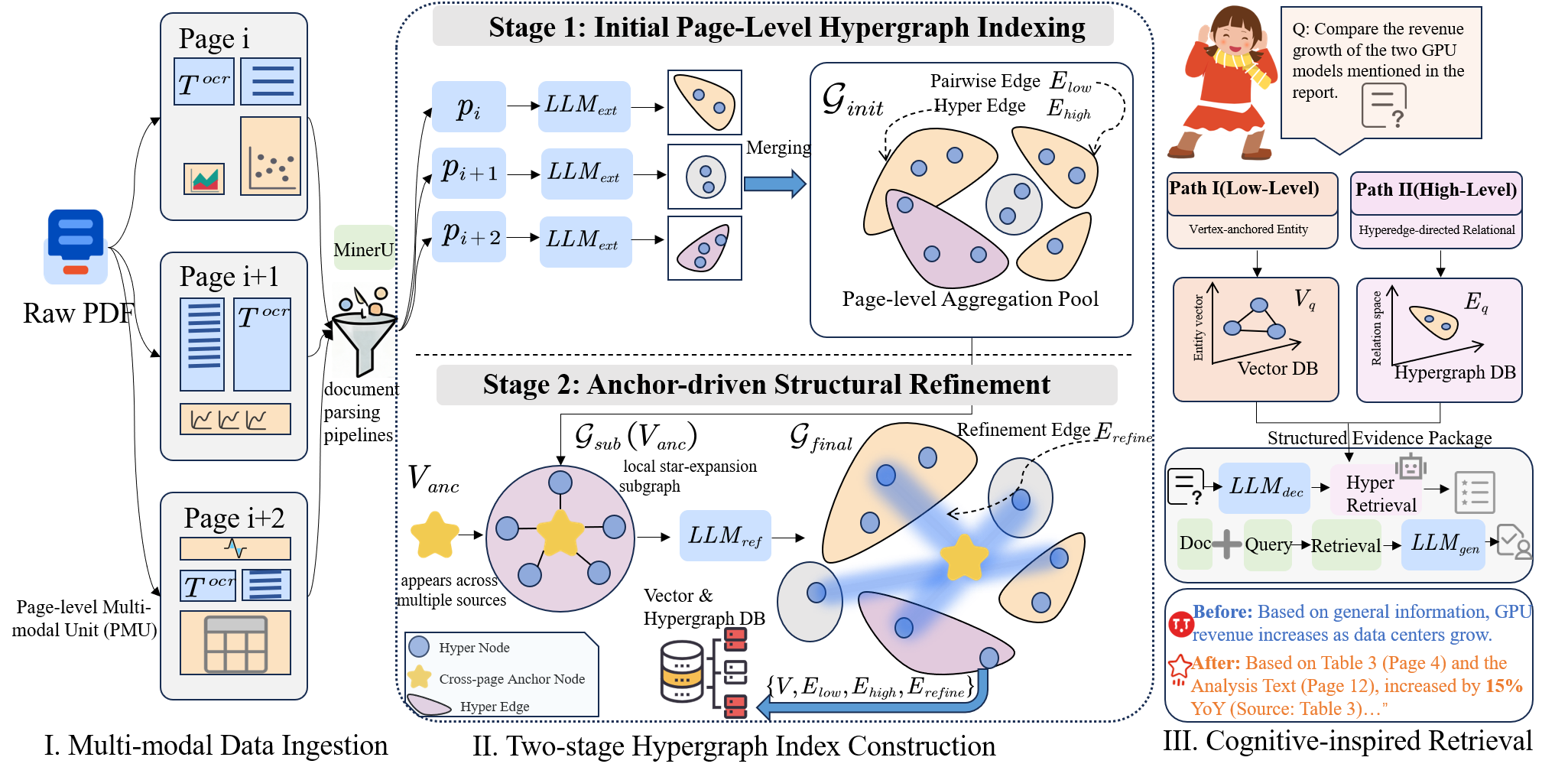}
    \caption{\textbf{Overview of the Hyper-M2RAG framework.} The system progresses through three core stages: (1) \textbf{Multi-modal Data Ingestion} via PMUs; (2) \textbf{Two-stage Hypergraph Construction} featuring anchor-driven structural refinement; (3) \textbf{Cognitive-inspired Hybrid Retrieval}. The comparative case (bottom right) highlights our advantage in cross-page factual grounding over vanilla LLMs.}
    \label{fig:main_framework}
\end{figure*}

\section{Related Work}
Traditional RAG typically relies on flattened chunk retrieval, which struggles to capture long-range dependencies. To bridge this gap, graph-enhanced approaches like GraphRAG \cite{edge2024local} and LightRAG \cite{guo2024lightrag} extract entity-centric relations to improve multi-hop reasoning. However, these binary graph models are limited to pairwise associations. To model higher-order group semantics, Hypergraph-based RAG (HyperRAG) \cite{feng2025hyper} introduces hyperedges to connect multiple vertices simultaneously. Recent variants like CogRAG\cite{hu2026cog} further explore cognitive-inspired hypergraph structures to enhance adaptive reasoning. While these methods excel in text-based high-order modeling, they remain largely "vision-blind" when faced with complex multi-modal layouts.

Recent studies have begun extending RAG to multi-modal domains, encompassing cross-modal alignment, specialized document parsing, and multi-source retrieval strategies \cite{zhang2025comprehensive}. Systems such as RAG-Anything \cite{guo2025rag}, MegaRAG \cite{hsiao2025megarag}, and MHier-RAG \cite{gong2025mhier} attempt to retrieve interleaved visual and textual components from complex layouts, while complementary approaches address query reformulation \cite{li2024dmqr}, heterogeneous source aggregation \cite{liu2025hm}, multi-turn reasoning \cite{joshi2024robust}, and evidence fusion \cite{yang2024rag}. However, existing methods suffer from retrieval-modality decoupling: they either treat images as secondary "post-retrieval supplements" or rely on lossy text-only indices (e.g., captions) that collapse the original structural layout. This leads to critical information decay and visual blind spots, especially when evidence spans across multiple pages. Unlike these approaches, our work pursues a native multi-modal indexing paradigm that preserves high-order structural and visual integrity within a unified hypergraph.

\section{Task Description and Preliminaries}

The objective of multi-modal long-context Document-based Question Answering (Doc-QA) is to generate a comprehensive answer $\mathcal{Y}_{ans}$ based on a query $\mathcal{Q}$ and a document $\mathcal{D}$ containing heterogeneous modalities. We formalize the generation process as:
\begin{equation}
Y_{ans} = \text{LLM}_{gen} (y_i \mid y_{<i}, Q, S),
\end{equation}
where $S$ represents the Structured Evidence Package retrieved from the document-indexed corpus. 

In our \textbf{Hyper-M2RAG} framework, the document $D$ is organized into a multi-modal hypergraph $G = (V, E)$, where $V$ denotes a set of normalized semantic entities vertices and $E = \{E_{low} \cup E_{high} \cup E_{refine}\}$ represents the set of hyperedges capturing multi-order relationships. Specifically, $E_{low}$ and $E_{high}$ capture intra-page pairwise and group associations, respectively, while $E_{refine}$ encapsulates latent cross-page dependencies reconstructed through anchor-driven refinement. 
The retrieval component is defined as a mapping function:
\begin{equation}
S = \text{HyperRetriever}(G, Q),
\end{equation}
where $\text{HyperRetriever}(\cdot)$ executes a dual-path alignment across the symbolic hypergraph and the vector space. This strategy ensures that $S$ integrates both vertex-anchored entity evidence $C_{ent}$ and hyperedge-directed relational context $C_{rel}$, facilitating complex reasoning over the high-order multi-modal substrate.

\section{Methodology}
\label{sec:method}

The core workflow of Hyper-M2RAG is illustrated in Fig.~\ref{fig:main_framework}, highlighting the transition from fragmented multi-modal pages to a unified, high-order hypergraph index.
\subsection{Multi-modal Data Ingestion}
To better preserve heterogeneous document structures, we redefine the primary unit of data ingestion for structured knowledge extraction. Unlike conventional RAG systems that treat documents as linear text streams, we adopt the Page-level Multi-modal Unit (PMU) as the fundamental atomic unit. Specifically, we utilize MinerU as a high-fidelity engine to parse raw PDFs into a structured sequence $\mathcal{D} = \{p_1, p_2, \dots, p_n\}$. The $i$-th unit $p_i$ is defined as:
\begin{equation}
p_i = \{ T_i^{ocr}, I_i^{page}, \mathcal{F}_{i,local} \},
\end{equation}
where $T_i^{ocr}$ is the aggregated textual content, $I_i^{page}$ is the global page image, and $\mathcal{F}_{i, local} = \{ I_{i,j}^{fig} \}_{j=1}^{n_i}$ represents localized image patches (e.g., figures, tables). 
This representation assigns distinct semantic roles to visual inputs: $I_i^{page}$ provides macro-visual context for layout awareness, while $\mathcal{F}_{i,local}$ offers fine-grained evidence from detected regions. To ensure robust ingestion, vision-dominant pages with sparse text are preserved via semantic placeholders (e.g., \textit{``Please see the Figures''}). Furthermore, this unified schema treats visual fields as optional ($\emptyset$), allowing the system to handle both text-only and multi-modal data through a consistent interface.

\subsection{Multi-modal Hypergraph Index Construction}

The construction of the HyperIndex follows a hierarchical paradigm: \textbf{Initial Page-level Extraction} followed by \textbf{Anchor-driven Structural Refinement}. This process transforms fragmented page-level units into a cohesive, high-fidelity semantic network.

\subsubsection{Initial Page-level Indexing and Aggregation}
For each PMU $p_i \in \mathcal{D}$, we first execute a multi-modal extraction operator $\text{LLM}_{ext}$ to capture the localized entity set  $V_i$ and their corresponding high-order associations $E_i$. Unlike traditional text-based extraction, this process simultaneously digests the textual substrate and hierarchical visual evidence. The extraction is formalized as follows:
\begin{equation}
\left\{
\begin{aligned}
V_i &= \text{LLM}_{ext}(P_{ent}(p_i)) \\
E_{i,low} &= \text{LLM}_{ext}(P_{low}(p_i, V_i)) \\
E_{i,high} &= \text{LLM}_{ext}(P_{high}(p_i, V_i)) 
\end{aligned}
\right. \quad \text{for } p_i \in \mathcal{D},
\end{equation}
where $P_{ent}$, $P_{low}$, and $P_{high}$ are specialized prompts (detailed in Appendix) for identifying vertices, pairwise relations, and high-order correlations, respectively. Each extracted element is tagged with a provenance attribute $\text{src} = \{i\}$ to anchor semantic entities to their original page index $i$.

To synthesize a document-level structure, a Canonical Merging operator $\mathcal{M}$ is applied to aggregate page-level sets. Vertices are normalized via case-insensitive alignment, while hyperedges are collapsed based on their unique, sorted constituent vertex sets:
\begin{equation}
\mathcal{G}_{init} = \mathcal{M} \left( \bigcup_{i=1}^n \{V_i, E_{i,low}, E_{i,high}\} \right),
\end{equation}
where $\mathcal{G}_{init}$ denotes the initial global hypergraph. In this stage, cross-page alignment is achieved through entity unification, where each global node or edge maintains an aggregated provenance index $\text{src}_{global} = \bigcup \text{src}_i$.

\subsubsection{Anchor-driven Structural Refinement}

While $\mathcal{G}_{init}$ aggregates local evidence, it remains constrained by the page-level receptive field, often failing to capture long-range semantic dependencies that transcend individual page boundaries. To bridge these disparate fragments, we introduce a refinement stage centered on Cross-page Anchors $V_{anc}$. These anchors are identified as entities with high provenance frequency, representing consistent semantic pivots across the document:
\begin{equation}
V_{anc} = \{ v \in V \mid |\text{src}(v)| \geq \tau \},
\end{equation}
where $\tau$ is the frequency threshold. 

Instead of re-processing entire page sequences with high computational redundancy as in MegaRAG, our approach performs targeted refinement within the topological neighborhoods of $V_{anc}$. 

%For each anchor $v \in \mathcal{V}_{anc}$, the system extracts its 1-hop structural neighborhood $\mathcal{G}_{sub}(v)$ from the hypergraph. 
For each anchor $v \in V_{anc}$, the system extracts its star-expansion subgraph $G_{sub}(v) = (V_{sub}, E_{sub})$ to encapsulate the multi-source contexts linked to this pivot through shared hyperedges:
\begin{equation}
\begin{cases}
E_{sub}(v) = \{ e \in (E_{low} \cup E_{high}) \mid v \in e \} \\
V_{sub}(v) = \{ u \in V \mid \exists e \in E_{sub}(v), u \in e \}
\end{cases}
\end{equation}
where $E_{sub}$ represents the set of all hyperedges (both pairwise and high-order) incident to $v$, and $V_{sub}$ denotes the union of all vertices contained within these incident hyperedges. 

This localized sub-hypergraph $G_{sub}$ acts as a structural nexus, providing a condensed reasoning substrate for the refinement operator $\mathcal{R}$ to bridge disparate page-level fragments and synthesize latent cross-page associations:
\begin{equation}
E_{refine} = \text{LLM}_{ref}(\mathcal{R}(G_{sub}(v), P_{ref})),
\end{equation}
where $P_{ref}$ is a specialized prompt directing the model to synthesize novel associations or consolidate existing relations by reconciling multi-source evidence. Finally, the refined relations are integrated into the global multi-modal hypergraph $\mathcal{G}_{final}$:
\begin{equation}
\mathcal{G}_{final} = \{V, E_{low}, E_{high}, E_{refine}\}.
\end{equation}

By shifting the refinement focus from raw page sequences to topological anchor neighborhoods, the system effectively eliminates the redundant processing of stable local information. This ensures a more precise discovery of high-order relational structures while maintaining significant computational efficiency.

\subsection{Cognitive-inspired Hybrid Retrieval}

Guided by the hierarchical structure of the \textit{HyperIndex}, we design a cognitive-inspired hybrid retrieval strategy to bridge user queries with high-fidelity evidence. As illustrated in our framework, the process initiates with a Query Decomposition stage, followed by a Dual-channel Retrieval mechanism that traverses both the symbolic hypergraph and the dense vector space.

\subsubsection{Multi-modal Query Decomposition}
For a given user query $q$, the system first employs an MLLM-based decomposition operator to extract two levels of semantic anchors:
\begin{equation}
\{ K_{low}, K_{high} \} = \text{LLM}_{dec}(P_{dec}(q)),
\end{equation}
where $K_{low}$ represents fine-grained entity keywords (e.g., specific objects or technical terms), and $K_{high}$ denotes high-level thematic keywords (e.g., abstract concepts or cross-page relations). $P_{dec}$ is the decomposition prompt detailed in the Appendix.

\subsubsection{Dual-channel Evidence Retrieval}
The retrieval process simultaneously executes two complementary paths to collect structured evidence from the HyperIndex:
\paragraph{Path I: Vertex-anchored Entity Retrieval} 
This path targets fine-grained evidence by anchoring the query to specific semantic vertices within the hypergraph. For each keyword $k \in K_{low}$, the system identifies the most relevant vertices $V_{q}$ via semantic similarity search in the entity vector space:
\begin{equation}
V_{q} = \{ v \in V \mid \text{sim}(\phi(k), \phi(v)) > \gamma \},
\end{equation}
where $\phi(\cdot)$ denotes the embedding function and $\gamma$ is the similarity threshold. Using $V_{q}$ as entry points, the entity-centric context $C_{ent}$ is expanded via their incident hyperedges in the HypergraphDB:
\begin{equation}
C_{ent} = \{ e \in (E_{low} \cup E_{high} \cup E_{refine}) \mid \exists v \in V_{q}, v \in e \}.
\end{equation}
This operation captures the topological neighbors of the query entities as structural evidence, preserving the immediate relational context and leveraging refined cross-page associations.

\paragraph{Path II: Hyperedge-directed Relational Retrieval} 
To capture overarching themes and cross-page narratives, this path interfaces directly with the relational vector index using $K_{high}$. For each thematic keyword $k \in K_{high}$, the system retrieves a candidate set of high-order hyperedges $E_{q}$ whose semantic embeddings align with the query intent:
\begin{equation}
E_{q} = \{ e \in (E_{low} \cup E_{high} \cup E_{refine}) \mid \text{sim}(\phi(k), \phi(e)) > \gamma \},
\end{equation}
where $E_{q}$ specifically targets the refined associations that transcend individual page boundaries. The relation-centric context $C_{rel}$ is then synthesized by back-tracing the constituent vertices and their multi-source provenance:
\begin{equation}
C_{rel} = \{ \text{src}(e), \text{nodes}(e) \mid e \in E_{q} \},
\end{equation}
where $\text{nodes}(e)$ extracts the semantic entities within the hyperedge and $\text{src}(e)$ retrieves the original multi-page text chunks. This path ensures the model reconciles broad conceptual queries with the high-order structural evidence consolidated during the refinement stage.

\subsubsection{Evidence Fusion and Generation}

Finally, the system integrates the multi-source evidence from both retrieval channels into a unified, structured context $S = \text{Fuse}(C_{ent}, C_{rel})$. Unlike conventional RAG systems that return fragmented raw text, $S$ is a Structured Evidence Package comprising: (i) normalized entities, (ii) high-order hyperedges, and (iii) their synchronized multi-page source chunks. The final response $R$ is synthesized as:
\begin{equation}
R = \text{LLM}_{gen}(P_{gen}(q, S)),
\end{equation}
where $P_{gen}$ is a generation prompt that instructs the model to reason over the provided topological structures and cross-page evidence.

The efficacy of this dual-channel paradigm is fundamentally rooted in the structural depth of the $HyperIndex$. Specifically, the cross-page hyperedges $E_{refine}$ generated during the anchor-driven refinement stage enable Path II to retrieve integrated evidence that transcends individual page boundaries. By reconciling these high-order associations with local entity-centric details, the framework provides a holistic and factually grounded understanding of complex, long-form documents.

\section{Experiments}
\label{sec:experiments}

In this section, we provide a comprehensive evaluation of Hyper-M2RAG. We first detail our experimental configurations and benchmark selections, followed by a quantitative analysis comparing our approach against state-of-the-art RAG baselines.

\subsection{Datasets}

To evaluate Hyper-M2RAG’s proficiency in managing multi-modal document structures encompassing both holistic understanding and granular retrieval, we conduct experiments across two task granularities: Global QA and Local QA.

\subsubsection{Global QA}
To evaluate the system's ability to synthesize information across entire document collections (book-level), we employ both textual and multi-modal benchmarks:

\begin{itemize}[leftmargin=*, noitemsep, topsep=2pt]
    \item \textbf{Textual Corpus:} We utilize the \textit{Mixed-Domain} ($\approx$0.62M tokens) from the \textbf{UltraDomain} \cite{qian2025memorag} dataset and  \textit{NeurologyCorp} ($\approx$1.97M tokens) from MedRAG \cite{xiong2024benchmarking}. These corpora specifically test the model's capacity for maintaining long-range semantic consistency within dense, specialized knowledge domains.
    \item \textbf{multi-modal Benchmark:} To address the scarcity of standardized multi-modal global QA datasets, we utilize a specialized evaluation set characterized by dense, visually-rich content. This benchmark incorporates a World History textbook (788 pages) and a Sustainable Development Report (SDR) slide deck (491 pages), both of which feature a high concentration of interleaved graphical elements, including intricate charts, tables, and thematic illustrations that are deeply integrated with the primary text.
\end{itemize}
\subsubsection{Global Question Generation Protocol}
To address the lack of human-labeled queries, we use an automated protocol with two modality-specific pipelines:
\begin{itemize}[leftmargin=*, noitemsep, topsep=2pt]
    \item \textbf{Textual Generation:} Using the document outline as a scaffold, we prompt an LLM to simulate 5 distinct professional users. For each user, we define 5 strategic tasks, each generating 5 complex questions that require a holistic understanding of the document. This results in 125 global questions per dataset.

    \item \textbf{multi-modal Generation:} For vision-heavy documents, we establish a generation pipeline. We sample 25 representative pages per dataset (organized into 5 batches of 5 pages). Each batch involves a specific user persona and 5 tasks. Crucially, we enforce a multi-modal-dependency constraint: questions must be unanswerable by text alone, requiring reasoning over visual data such as map coordinates, chart trends, or diagram structures. This ensures 125 high-quality multi-modal questions per dataset.
\end{itemize}

\subsubsection{Local QA}
To evaluate retrieval precision at the granular level (page- or slide-level), we utilize the RealMMBench \cite{wasserman2025real} benchmark. RealMMBench is specifically designed to assess multi-modal RAG systems under stressed conditions, including visual-rich layouts, table-heavy content, and sophisticated query rephrasing.
While RealMMBench spans multiple domains, we perform a strategic evaluation on the TechReport (1,674 pages) and TechSlides (1,963 pages) sub-datasets. These technical domains are characterized by dense high-order structural dependencies, making them ideal for validating our high-order hypergraph construction and anchor-driven refinement mechanisms. This selection ensures a rigorous stress test of document intelligence while maintaining computational feasibility during the evaluation process.

\subsection{Baselines and Evaluation Metrics}
We compare \textbf{Hyper-M2RAG} against several state-of-the-art baselines: GraphRAG \cite{edge2024local}, a widely adopted knowledge graph framework; HyperRAG \cite{feng2025hyper}, which utilizes text-only hypergraphs; and MegaRAG \cite{hsiao2025megarag}, a multi-modal graph-based RAG system. To ensure a fair assessment, we evaluate all methods across both textual and multi-modal benchmarks to test their core retrieval and reasoning capabilities.

\subsubsection{Global QA Metrics}
Given the absence of ground-truth answers for book-level queries, prior works often rely on LLM-based evaluation. However, such evaluations are frequently prone to \textit{position bias}, where the evaluator model favors responses presented earlier in the prompt. To mitigate this and ensure reliability, we implement a Dual-Response Swap \& Average strategy:
\begin{itemize}[leftmargin=*, noitemsep, topsep=2pt]
    \item \textbf{Dual-Prompt Evaluation:} For every pair of responses $(A, B)$, we generate two distinct evaluation prompts: a forward prompt (A followed by B) and a reversed prompt (B followed by A).
    \item \textbf{Result Normalization:} To maintain a consistent reference frame, results from the reversed evaluation are flipped (e.g., if B was preferred in the reversed prompt, it is mapped to a preference for the second position in the forward frame).
    \item \textbf{Consistency Arbitration:} We compare the forward and normalized reversed results. If the evaluator remains consistent across both prompts, the result is recorded. If the two evaluations conflict, the pair is marked as a \textbf{Tie} to filter out stochastic noise and ensure robust win-rate calculation.
\end{itemize}

Responses are assessed across four qualitative dimensions following \cite{guo2024lightrag}: (1) Comprehensiveness: Coverage of all query facets; (2) Diversity: Richness of perspectives; (3) Empowerment: Support for user understanding; and (4) Overall: An aggregate measure of the preceding criteria. 
\begin{table*}[t]
\centering
\caption{Win-rate analysis comparing \textbf{Hyper-M2RAG} against GraphRAG, HyperRAG, and MegaRAG across four distinct domains. Results are presented as percentages (\%). Bold values indicate the winning method (excluding ties).}
\label{tab:main_results}
\small
\begin{tabular}{@{}l ccc @{\hskip 0.3in} ccc @{\hskip 0.3in} ccc @{\hskip 0.3in} ccc@{}}
\toprule
\multirow{2}{*}{\textbf{Metrics}} & \multicolumn{3}{c}{\textbf{Mix}} & \multicolumn{3}{c}{\textbf{Neurology}} & \multicolumn{3}{c}{\textbf{World History}} & \multicolumn{3}{c}{\textbf{Sustainable Report}} \\
\cmidrule(lr){2-4} \cmidrule(lr){5-7} \cmidrule(lr){8-10} \cmidrule(lr){11-13}
 & GraphRAG & \textbf{Ours} & Tie & GraphRAG & \textbf{Ours} & Tie & GraphRAG & \textbf{Ours} & Tie & GraphRAG & \textbf{Ours} & Tie \\
\midrule
Comprehensiveness & 0.0 & \textbf{46.4} & 53.6 & 0.8 & \textbf{37.6} & 61.6 & 0.8 & \textbf{47.2} & 52.0 & 0.0 & \textbf{62.4} & 37.6 \\
Diversity         & 11.2 & \textbf{60.8} & 28.0 & 9.6 & \textbf{56.0} & 34.4 & 4.8 & \textbf{69.6} & 25.6 & 3.2 & \textbf{91.2} & 5.6 \\
Empowerment       & 4.0 & \textbf{56.0} & 40.0 & 3.2 & \textbf{61.6} & 35.2 & 5.6 & \textbf{66.4} & 28.0 & 1.6 & \textbf{77.6} & 20.8 \\
\rowcolor{gray!10} Overall & 2.4 & \textbf{51.2} & 46.4 & 2.4 & \textbf{52.0} & 45.6 & 4.8 & \textbf{61.6} & 33.6 & 0.8 & \textbf{77.6} & 21.6 \\
\midrule
\midrule
 & HyperRAG & \textbf{Ours} & Tie & HyperRAG & \textbf{Ours} & Tie & HyperRAG & \textbf{Ours} & Tie & HyperRAG & \textbf{Ours} & Tie \\
\cmidrule(lr){2-4} \cmidrule(lr){5-7} \cmidrule(lr){8-10} \cmidrule(lr){11-13}
Comprehensiveness & 0.0 & \textbf{8.8} & 91.2 & 6.4 & \textbf{17.6} & 76.0 & 0.0 & \textbf{44.0} & 56.0 & 1.6 & \textbf{29.6} & 68.8 \\
Diversity         & 23.2 & \textbf{35.2} & 41.6 & 3.2 & \textbf{18.4} & 78.4 & 15.2 & \textbf{54.4} & 30.4 & 17.6 & \textbf{43.2} & 39.2 \\
Empowerment       & 15.2 & \textbf{28.8} & 56.0 & 14.4 & \textbf{28.8} & 56.8 & 16.8 & \textbf{46.4} & 36.8 & 20.8 & \textbf{26.4} & 52.8 \\
\rowcolor{gray!10} Overall & 13.6 & \textbf{27.2} & 59.2 & 11.2 & \textbf{22.4} & 66.4 & 14.4 & \textbf{44.8} & 40.8 & 11.2 & \textbf{25.6} & 63.2 \\
\midrule
\midrule
 & MegaRAG & \textbf{Ours} & Tie & MegaRAG & \textbf{Ours} & Tie & MegaRAG & \textbf{Ours} & Tie & MegaRAG & \textbf{Ours} & Tie \\
\cmidrule(lr){2-4} \cmidrule(lr){5-7} \cmidrule(lr){8-10} \cmidrule(lr){11-13}
Comprehensiveness & 4.8 & \textbf{53.6} & 41.6 & 0.0 & \textbf{20.0} & 80.0 & 0.0 & \textbf{44.0} & 56.0 & 0.0 & \textbf{64.0} & 36.0 \\
Diversity         & 8.8 & \textbf{44.8} & 46.4 & 6.4 & \textbf{52.8} & 40.8 & 2.4 & \textbf{75.2} & 22.4 & 1.6 & \textbf{90.4} & 8.0 \\
Empowerment       & 3.2 & \textbf{63.2} & 33.6 & 0.8 & \textbf{44.8} & 54.4 & 3.2 & \textbf{62.4} & 34.4 & 1.6 & \textbf{90.4} & 8.0 \\
\rowcolor{gray!10} Overall & 5.6 & \textbf{58.4} & 36.0 & 0.0 & \textbf{36.8} & 63.2 & 1.6 & \textbf{59.2} & 39.2 & 1.6 & \textbf{88.8} & 9.6 \\
\bottomrule
\end{tabular}
\end{table*}
\subsubsection{Local QA Metrics}
For granular (page- or slide-level) QA, performance is measured by semantic alignment with ground-truth reference answers. We utilize a powerful LLM as an automated judge to determine semantic consistency between the generated and reference answers, reporting Accuracy as the primary metric.

\subsection{Implementation Details}

To ensure consistency across all evaluated RAG frameworks, we standardize the backbone models and processing pipeline. For response generation, we utilize Qwen3-VL-8B \cite{bai2025qwen3}, while DeepSeek-v3 \cite{liu2024deepseek} is employed for synthetic question generation and evaluation due to its superior reasoning robustness. 

All methods share a unified embedding space facilitated by gme-Qwen2-VL-2B-Instruct \cite{zhang2024gme}, which supports single-, cross-, and fused-modality retrieval tasks. Textual corpora are partitioned into 1,200-token chunks with a 100-token overlap. For multi-modal documents, we leverage the MinerU2.5 toolkit \cite{niu2025mineru2} to extract text, figures, and tables. MinerU's ability to preserve complex layouts and symbols in machine-readable formats is particularly effective for the technical documents used in our study.

\subsection{Global QA Performance}

The primary results in Table \ref{tab:main_results} compare Hyper-M2RAG against three state-of-the-art baselines across four domains: two purely textual corpora (Mix and Neurology) and two multimodal datasets (World History and Sustainable Report). While our method maintains a consistent lead in textual tasks, it achieves a decisive, overwhelming victory in multimodal scenarios. Our analysis yields the following key insights:

\paragraph{Superiority across Diverse Domains} 
\textbf{Hyper-M2RAG} consistently outperforms GraphRAG, HyperRAG, and MegaRAG across all qualitative dimensions. Notably, in the Sustainable Report domain characterized by high-density multimodal content and complex layouts, our method achieves a dominant Overall win rate of 88.8\% against MegaRAG and 77.6\% against GraphRAG. These results demonstrate that our high-order hypergraph structure captures multi-scale semantic dependencies in technical documents more effectively than traditional binary graphs.

\paragraph{Gains in Diversity and Empowerment}
Hyper-M2RAG demonstrates a substantial lead in Diversity and Empowerment metrics. For instance, in the World History domain, our method achieves a 75.2\% win rate in Diversity against MegaRAG. This performance gain is driven by our hyper-relational modeling mechanism, which transcends the limitations of binary-edge retrieval. Unlike traditional graphs that only link pairwise entities, our high-order hyperedges group multi-modal entities into unified semantic clusters. This allows the retriever to capture non-local, multi-hop connections across disparate document sections in a single traversal. 
%By aggregating these exhaustive and multifaceted perspectives, Hyper-M2RAG provides more comprehensive insights, leading to significantly higher Empowerment scores.

\paragraph{Robustness against Hypergraph Baselines}
Compared to the text-only HyperRAG, Hyper-M2RAG maintains a consistent margin of improvement, particularly in Overall win rates ranging from 22.4\% to 44.8\%. While HyperRAG introduces basic hyper-structures, our anchor-driven refinement and multimodal integration ensure that the constructed hypergraphs are semantically precise rather than merely dense. The high tie rates in Comprehensiveness (e.g., 91.2\% in Mix) indicate that while most hypergraph methods achieve basic content coverage, Hyper-M2RAG excels in the qualitative depth and organizational clarity of the generated responses.

\paragraph{Multimodal Advantage}
In multimodal datasets like World History and Sustainable Report, the performance gap between Hyper-M2RAG and baselines widens significantly. This advantage stems from our layout-aware hyperedges, which directly link visual elements to their corresponding textual descriptions within a unified topological space. For instance, the 90.4\% Empowerment score in the Sustainable domain demonstrates the model's superior ability to interpret and synthesize information from complex tables and figures.

\newcolumntype{L}[1]{>{\raggedright\arraybackslash}p{#1}}

% --- 表格 1：Local QA ---
\begin{table}[h]
\centering
\small % 统一字号
\caption{Local QA performance (Accuracy \%) on RealMMBench.}
\label{tab:realmmbench_local_qa}
\begin{tabular}{@{}L{1.6in} cc@{}} % 手动指定第一列宽度为 1.6 英寸
\toprule
\textbf{Method} & \textbf{TechReport} & \textbf{TechSlides} \\
\midrule
GraphRAG & 30.0 & 57.0 \\
HyperRAG & 56.0 & 67.0 \\
MegaRAG  & 68.0 & 69.0 \\
\midrule
\rowcolor{gray!10} \textbf{Hyper-M2RAG (Ours)} & \textbf{70.0} & \textbf{78.0} \\
\bottomrule
\end{tabular}
\end{table}
\subsection{Local QA Performance}

Table \ref{tab:realmmbench_local_qa} reports the accuracy of different methods on the TechReport and TechSlides sub-datasets of RealMMBench. Hyper-M2RAG achieves superior performance, reaching 70.0\% and 78.0\% accuracy, respectively. 
Compared with the state-of-the-art baseline MegaRAG, our method achieves a notable 9 percentage point improvement on TechSlides. This specific gain highlights the efficacy of our framework in visual-heavy contexts. Technical slides often contain fragmented information across different visual blocks; while MegaRAG relies on dense graph connectivity, our high-order hypergraph structure more effectively clusters spatially disparate but semantically related elements (e.g., a figure and its corresponding bullet points), leading to more precise local retrieval.

\subsection{Ablation Study}

To investigate the contribution of different components in Hyper-M2RAG, 
we progressively introduce multi-modal extraction and anchor-driven refinement 
based on the HyperRAG baseline, as shown in Table~\ref{tab:ablation_study}. 
Our analysis yields two primary insights:

\paragraph{Effect of Multi-modal Hypergraph Construction}
Adding multi-modal extraction (\textit{+ MM Extraction}) consistently improves 
performance over the text-only HyperRAG baseline on both TechReport and 
TechSlides datasets. This demonstrates that incorporating visual elements and 
layout information helps preserve cross-modal semantic dependencies, enabling 
more complete evidence retrieval from complex documents.

\paragraph{Effect of Anchor-driven Refinement}
Introducing anchor-driven refinement further improves the performance over the 
multi-modal hypergraph variant. This verifies that selectively reconstructing 
local neighborhoods around cross-page anchors effectively captures long-range 
dependencies while avoiding unnecessary processing of irrelevant regions.

Overall, the ablation results confirm that both multi-modal representation 
and anchor-driven refinement are essential components of Hyper-M2RAG, with 
their combination enabling more effective multimodal document understanding.

\begin{table}[h]
\centering
\small % 统一字号，与上表绝对一致
\caption{Ablation study of Hyper-M2RAG components.}
\label{tab:ablation_study}
\begin{tabular}{@{}L{1.6in} cc@{}} % 同样指定 1.6 英寸，确保视觉对齐
\toprule
\textbf{Variant} & \textbf{TechReport} & \textbf{TechSlides} \\
\midrule
HyperRAG (Baseline) & 56.0 & 67.0 \\
\quad + MM Extraction & 64.0 & 74.0 \\
\quad + Anchor Refinement & 60.0 & 69.0 \\
\midrule
\rowcolor{gray!10} \textbf{Hyper-M2RAG (Full)} & \textbf{70.0} & \textbf{78.0} \\
\bottomrule
\end{tabular}
\end{table}

\begin{table*}[t] % 强制置顶
\centering
\caption{Comparison of structural hyperedge counts across datasets.} % 记得加Caption，否则容易被排版引擎忽略
\label{tab:counts}
\small % 如果空间不够，稍微缩小字体
\begin{tabularx}{\textwidth}{@{}l XXXXXX@{}} % 使用tabularx自动填满整行
\toprule
\multirow{2}{*}{\textbf{Variant}} & \multicolumn{3}{c}{\textbf{TechReport} (1,674 pages)} & \multicolumn{3}{c}{\textbf{TechSlides} (1,963 pages)} \\
\cmidrule(lr){2-4} \cmidrule(lr){5-7}
 & \textbf{Low} & \textbf{High} & \textbf{Total} & \textbf{Low} & \textbf{High} & \textbf{Total} \\
\midrule
HyperRAG (Baseline) & 12,957 & 2,579 & 15,536 & 8,543 & 1,716 & 10,259 \\
HyperRAG + MM       & 10,016 & 2,078 & 12,094 & 9,309 & 2,004 & 11,313 \\
\midrule
\rowcolor{gray!10} \textbf{Hyper-M2RAG (Ours)} & \textbf{14,221} & \textbf{8,257} & \textbf{22,478} & \textbf{12,522} & \textbf{6,571} & \textbf{19,093} \\
\bottomrule
\end{tabularx}
\end{table*}

% \begin{figure*}[t]
%   \centering
%   \includegraphics[width=0.98\textwidth]{p4.png}
%   \caption{Qualitative comparison of RAG responses on a cross-page multi-modal query. (a) \textbf{HyperRAG} suffers from "confident hallucination" due to missing visual grounding. (b) \textbf{MegaRAG} correctly retrieves fragments but fails to bridge the 88-page semantic gap. (c) Our \textbf{Hyper-M2RAG} achieves logical closure by linking causal text with visual UI elements across disparate sections.}
%   \label{fig:case_study}
% \end{figure*}

\subsection{Structural Complexity Analysis}

To investigate the structural characteristics of \textbf{Hyper-M2RAG}, Table~\ref{tab:counts} compares the hypergraph statistics of different methods, with particular focus on high-order relations (degree $>2$).

\paragraph{High-Order Relational Modeling}
\textbf{Hyper-M2RAG} constructs substantially more high-order hyperedges than existing approaches. On \textit{TechReport}, Hyper-M2RAG builds 8,257 high-order hyperedges, compared with 2,579 from HyperRAG, representing a \textbf{220\% increase}. Similarly, on \textit{TechSlides}, the number of high-order hyperedges increases from 1,716 to 6,571. Since high-order hyperedges explicitly represent dependencies among multiple entities, this increase indicates that Hyper-M2RAG provides a richer structural representation beyond conventional pairwise relations.

\paragraph{Effect of Multi-modal Hypergraph Construction}
Interestingly, simply incorporating multi-modal information does not necessarily lead to more high-order structures. The HyperRAG+MM variant produces fewer high-order hyperedges than HyperRAG on \textit{TechReport} (2,078 vs. 2,579). This suggests that modality augmentation alone is insufficient to establish effective high-order relations. In contrast, Hyper-M2RAG jointly models textual, visual, and layout elements through hypergraph construction and anchor-driven refinement, resulting in substantially richer high-order relational structures.

\paragraph{Overall Structural Connectivity}
Beyond high-order relations, Hyper-M2RAG also achieves the largest number of total hyperedges across both datasets, with 22,478 edges on \textit{TechReport} and 19,093 edges on \textit{TechSlides}. This demonstrates that our framework constructs a more densely connected document representation, providing a stronger structural foundation for retrieving evidence distributed across long and complex documents.

\begin{table}[h]
\centering
\small % 统一字号
\caption{ Multi-Judge Audit (\%). H-RAG = HyperRAG.}
\label{tab:judge}
\begin{tabular}{@{}l ccc @{\hskip 0.2in} ccc@{}}
\toprule
& \multicolumn{3}{c}{\textbf{Sustainable Report}} & \multicolumn{3}{c}{\textbf{World History}} \\
\cmidrule(lr){2-4} \cmidrule(lr){5-7}
\textbf{Evaluator} & H-RAG & \textbf{Ours} & Tie & H-RAG & \textbf{Ours} & Tie \\
\midrule
Kimi-K2.6      & 25.6 & \textbf{55.2} & 19.2 & 24.0 & \textbf{44.0} & 32.0 \\
GLM-5.1        & 25.6 & \textbf{51.2} & 23.2 & 24.8 & \textbf{38.4} & 36.8 \\
DeepSeek-V4 Pro & 15.2 & \textbf{41.6} & 43.2 & 20.0 & \textbf{36.0} & 44.0 \\
\bottomrule
\end{tabular}
\end{table}

\subsection{Cross-Model Judge Audit}
Table~\ref{tab:judge} presents a multi-judge evaluation on the Sustainable Report and World History corpora, conducted with three recently released, architecturally diverse LLMs: Kimi-K2.6, GLM-5.1, and DeepSeek-V4 Pro.
We select HyperRAG as the primary baseline for this audit because it is the most structurally aligned counterpart to Hyper-M2RAG: both methods organize knowledge using hypergraphs, while only Hyper-M2RAG introduces multimodal anchors and star-expansion refinement.
This design isolates the contribution of our multimodal mechanism from general hypergraph-induced gains.

Across both domains and all three evaluators, Hyper-M2RAG consistently outperforms HyperRAG, with win rates ranging from 36.0\% to 55.2\%.
Crucially, the preference for Hyper-M2RAG is not judge-specific: the same ranking trend holds for all model families, indicating that the observed improvements are not a single-judge artifact but a robust, cross-model phenomenon.

\begin{table}[h]
\centering
\small % 统一字号
\caption{\small Efficiency on World History (per 100 pages).}
\label{tab:eff}
\begin{tabular}{lccc}
\hline
\textbf{Metric} & \textbf{MegaRAG} & \textbf{HyperRAG + Page} & \textbf{Ours} \\ \hline
Build \& Refine Time (min) & 16 & 18 & 12 \\
Token Consumption ($10^3$) & 2,742.2 & 3,376.1 & 765.9 \\ \hline
\end{tabular}
\end{table}

\subsection{Reduction in Token Cost and Latency}
Table~\ref{tab:eff} provides a head-to-head efficiency comparison on the World History corpus (per 100 pages).
Hyper-M2RAG reduces build and refinement time to 12\,min and lowers token consumption by 72\% relative to MegaRAG and 77\% relative to HyperRAG+Page.
This gain stems from a fundamental shift in refinement granularity.

Rather than performing document- or page-level rescans during incremental updates, Hyper-M2RAG organizes multimodal evidence into a hypergraph, where anchors serve as structural pivots linking text, tables, figures, and layout elements.
Refinement is then confined to anchor-centered star neighborhoods: only hyperedges incident to updated anchors are recomputed.

Consequently, the update process is transformed from a global sweep into a topology-guided local operation.
This design eliminates redundant computation across unrelated content while preserving rich cross-modal linkages.
Because the number of anchors requiring updates grows more slowly than document length, the refinement cost scales sub-linearly, allowing Hyper-M2RAG to maintain high efficiency on long, multimodal documents without altering the input scope or representation.

\section{Conclusion}

In this paper, we presented \textbf{Hyper-M2RAG}, a novel multi-modal retrieval-augmented generation framework designed for complex, high-density technical documents. By transitioning from traditional binary graphs to a high-order hypergraph structure, our approach effectively captures the multi-scale semantic dependencies that are often fragmented across disparate document sections and modalities. 
Our core contribution lies in the integration of layout-aware hyperedge construction and an anchor-driven refinement mechanism. This synergy ensures that visual elements, such as tables and figures, are not merely treated as isolated inputs but are topologically aligned with their textual context within a unified semantic space. Furthermore, our dual-path retrieval strategy—comprising Vertex-anchored Entity Retrieval and Hyperedge-directed Relational Retrieval—enables the model to reconcile fine-grained evidence with overarching thematic narratives. 
Experimental results across four diverse domains demonstrate that Hyper-M2RAG consistently outperforms state-of-the-art baselines, particularly in multi-modal scenarios where it achieves significant gains in Diversity and Empowerment. 
% In this paper, we presented \textbf{Hyper-M2RAG}, a novel multi-modal retrieval-augmented generation framework for high-density technical documents. By transitioning to a high-order hypergraph structure, our approach captures multi-scale semantic dependencies across disparate sections and modalities that traditional binary graphs often fragment. 

% Our core contribution lies in the synergy between layout-aware hyperedge construction and anchor-driven refinement, ensuring visual elements are topologically aligned with textual context in a unified semantic space. This is complemented by a dual-path retrieval strategy—balancing vertex-anchored entities with hyperedge-directed narratives—to reconcile fine-grained evidence with global themes. Experimental results across four domains demonstrate that Hyper-M2RAG consistently outperforms state-of-the-art baselines, particularly in multi-modal scenarios requiring high diversity and empowerment. These findings establish hyper-relational modeling as a robust foundation for the next generation of multi-modal document intelligence.

\balance
\section{Acknowledgments}
This work was supported by Brain Science and Brain-like Intelligence Technology——National Science and Technology Major Project (2025ZD0217300), the National Key Research and Development Program of China under Grant (2023YFB4502803), the National Natural Science Foundation of China (No. U25A20532) and the Beijing Natural Science Foundation under Grant (No. L242167).

\bibliographystyle{ACM-Reference-Format}
%\bibliography{sample-base}
\bibliography{software}
\end{document}